%% file: Bilan_RFF_Nystrom.tex
\documentclass{article}
\usepackage{algorithm}
\usepackage{amssymb,amsmath}
\usepackage{graphicx}
\usepackage{url}

\newcommand{\argmin}{\operatornamewithlimits{argmin}}
\newtheorem{theorem}{Theorem}[section]
\newtheorem{proposition}[theorem]{Proposition}

\def\diag{{\rm diag}}
\def\dsp{\displaystyle}
\def\Dxi{\Delta x_i}

\def\Kh{\hat{K}}
\def\ks{k_\sigma}
\def\ksh{\hat{k}_\sigma}
\newcommand{\ls}[1]{\lambda_{\sigma,#1}} 
\def\Lsr{\Lambda_\sigma^{(r)}}
\def\Nys{Nystr\"{o}m }
\def\phish{\tilde{\phi}_\sigma}
\def\RR{\mathbb{R}}

\newcommand{\us}[1]{u_{\sigma,#1}}
\def\Usr{U_\sigma^{(r)}}
\def\xh{\hat{x}}

\def\bproof{\noindent{\bf Proof}:~}
\def\eproof{\nobreak\hfill\penalty80\hskip1em\hbox{}\nobreak\hfill$\Box$}

\begin{document}
\input{debut.tex}
\input{intro.tex}
\input{notations.tex}
\input{RoSVM.tex}
\input{NLRoSVM.tex}
\input{commentary}

\bibliographystyle{plain}
\bibliography{refs}

\end{document}

%% file: debut.tex
\title{Feature uncertainty bounding schemes\\ for large robust nonlinear SVM classifiers}

\author{Nicolas Couellan\thanks{Institut de Math\'{e}matiques de Toulouse, Universit\'{e} Paul Sabatier, 118 route de Narbonne 31062 Toulouse Cedex 9, France - nicolas.couellan@math.univ-toulouse.fr}         
\and
        Sophie Jan\thanks{Institut de Math\'{e}matiques de Toulouse, Universit\'{e} Paul Sabatier, 118 route de Narbonne 31062 Toulouse Cedex 9, France}
}

\date{}

\maketitle

\begin{abstract}
We consider the binary classification problem when data are large and subject to unknown but bounded uncertainties. We address the problem by formulating the nonlinear support vector machine training problem with robust optimization. To do so, we analyze and propose two bounding schemes for uncertainties associated to random approximate features in low dimensional spaces. The proposed techniques are based on Random Fourier Features and the \Nys methods. The resulting formulations can be solved with efficient stochastic approximation techniques such as stochastic (sub)-gradient, stochastic proximal gradient techniques or their variants.
\end{abstract}

%% file: intro.tex
\section{Introduction}

\noindent When designing classification models, the common idea is to consider that training samples are certain and not subject to noise. In reality, this assumption may be far from being verified. There are several sources of possible noise. For example, data may be noisy by nature. In biology, for instance, measurements on living cells under the miscroscope are unprecise due to the movement of the studied organism. Furthermore, any phenomenon that occurs at a smaller scale than the given microscope resolution is measured with errors. Uncertainty is sometimes also introduced as a way to handle missing data. If some of the sample attributes are not known but can be estimated, they can be replaced by their estimation, assuming some level of uncertainty. In the work that is presented here, we are interested in constructing optimization-based classification models that can account for uncertainty in the data and provide some robustness to noise. To do so, some knowledge on the noise is required. Most of the time, assuming that we know the probability distribution of the noisy perturbations is unrealistic. However, depending on the context, one may have knowledge on the mean and variance or simply the magnitude of the perturbations. The quantity and the nature of information available will be critical for the design of models that are immune to noise.

\noindent In binary classification problems, the idea is to compute a decision rule based on empirical data in order to predict the class on new observations. If empirical data is uncertain and no robustness is incorporated in the model, the decision rule will also lead to uncertain decision. In the context of Support Vector Machine (SVM) 
classification~\cite{vapnik1998}, there are several ways to construct robust counterparts of the classification problem, see for example~\cite{bental}. In these techniques, one considers that data points can be anywhere within an uncertainty set centered at a nominal point. The uncertainty set may have various shapes and dimension depending on the application and the level of noise. In the chance constrained SVM method, instead of ensuring that as many possible points are on the right side of the separating hyperplane, the idea is to ensure that a large probability of them will be correctly classified~\cite{BenTalBhadraBhattacharyyaNath,bental}. Alternatively, one may want to ensure that no matter where the data points are in the uncertainty set, the separating hyperplane will ensure that it will always be on the correct side. This is known as the worst case robust SVM approach~\cite{trafalis2,trafalisbook}. In both methods, one can show that the resulting classification problem can be formulated as a Second Order Cone Program (SOCP).

\noindent Classification problems are usually very large optimization problems due to the great number of samples available. Furthermore, so far, to the best of our knowledge, robust SVM formulations are only dealing with linear classification. If uncertainties are large, additional non-separability is introduced in the data that may limit the predictive performance of robust SVM models.

\noindent To propose solutions to these challenges, we present several new models and SVM classification approaches. We first describe the concept of robust Support Vector Machine (SVM) classification and its computation through stochastic optimization techniques when the number of samples is large. We further explain the concept of safe SVM we have introduced to overcome the conservatism issues in classical robust optimization techniques. We then develop new extensions to robust nonlinear classification models where we propose computational schemes to bound uncertainties in the feature space. 

%% file: notations.tex
\section{Notations}

We define the following notations:

\begin{itemize}
\item $\Sigma^{1/2}$ the matrix in $\RR^{n\times n}$ associated to the positive definite matrix $\Sigma$ such that $\Sigma^{1/2}(\Sigma^{1/2})^{\top}=\Sigma$ ;
\item $\Sigma^{-1/2}$ the matrix $(\Sigma^{1/2})^{-1}$ ;
\item $\Sigma^{\top/2}$ the matrix $(\Sigma^{1/2})^{\top}$ ;
\item $\mathcal{N}\left(\mu,\nu^2\right)$ the normal distribution with mean $\mu$ and variance $\nu^2$.
\end{itemize}

%% file: RoSVM.tex
\section{Robust SVM classification models: preliminaries}

\noindent We first recall the standard SVM 
methodology~\cite{vapnik1998,scholkopf2002,cristianini2000} to find the maximum margin separating hyperplane between two classes of data points.

\noindent Let $x_{i}\in \mathbb{R}^n$ be a collection of input training vectors for $i=1,...,L$ and $y_{i}\in\{-1,1\}$ be their corresponding labels. The SVM classification problem to separate the training vectors into the $-1$ and $1$ classes can be formulated as follows:
\begin{eqnarray}\label{primsvm}
\dsp \min_{w, b, \xi} & \frac{\lambda}{2} \|w\|^{2}+\dsp \sum_{i=1}^{L}\xi_{i}\\
\nonumber \mbox{subject to}
& y_{i}(\langle w,x_{i}\rangle +b)+\xi_{i} \geq 1  & i=1,\ldots,L\\
\nonumber & \xi_{i} \geq 0 & i=1,\ldots,L \\
\nonumber \end{eqnarray}
where $\xi_{i}$ are slack variables added to the constraints and in the objective as a penalty term, $\lambda$ is a problem specific constant controlling the trade-off between margin (generalization) and classification, and $\langle w,x\rangle+b=0$ defines the separating hyperplane with $w$ the normal vector to the separating hyperplane and $b$ its relative position to the origin.

\noindent Consider now instead a set of noisy training vectors $\left\{\tilde{x}_i \in \mathbb R^{n}, i=1,\ldots,L \right\}$ where $\tilde{x}_i=x_i+\Delta x_i$ for all $i$ in $\{1,...,L\}$ and $\Delta x_i$ is a random perturbation, Problem~(\ref{primsvm}) becomes:
\begin{eqnarray}
\label{noisysvm}
\dsp \min_{w, b, \xi} 
& \frac{\lambda}{2} \|w\|^{2}+\dsp \sum_{i=1}^{L}\xi_{i}
\\
\mbox{subject to}
& y_{i}(\langle w,x_i+\Delta x_i \rangle+b)+\xi_{i} \geq 1,
& \forall i=1,\ldots,L
\nonumber
\\
& \xi_{i} \geq 0,
& \forall i=1,\ldots,L.
\nonumber
\end{eqnarray}

\noindent Observe that the first constraint involves the random variable $\Delta x_i$ and Problem~(\ref{noisysvm}) cannot be solved as such. Extra knowledge on the perturbations is needed to transform (\ref{noisysvm}) into a deterministic and numerically solvable problem. If the uncertainty set is bounded and the bound is known, one can consider that 
$$\|\Sigma_i^{-1/2}\Delta x_i\|_{p}\leq \gamma_i, 
\quad\mbox{with } \gamma_i>0, \forall i\in\{1,\ldots,L\} 
\mbox{ and } p\geq 1.
$$
Various choices of $\Sigma_i$ and $p$ will lead to various types of uncertainties such as for example box-shaped uncertainty ($\parallel \tilde{x}_i - x_i \parallel_{\infty}\leq \gamma_i $), spherical uncertainty ($\parallel \tilde{x}_i- x_i \parallel_{2}\leq \gamma_i$) or ellipsoidal uncertainty ($(\tilde{x}_i- x_i)^{\top}\Sigma_i^{-1}(\tilde{x}_i- x_i)\leq \gamma_i^2$ for some positive definite matrix $\Sigma_i$).

\noindent To design a robust model, one has to satisfy the linear inequality constraint in Problem (\ref{noisysvm}) for every realizations of $\Delta x_i$. This can be done by ensuring the constraint in the worst case scenario for $\Delta x_i$, leading to the following robust counterpart optimization problem:
\begin{eqnarray}
\label{ronoisySVM}
\dsp\min_{w,b,\xi} \quad   
& \frac{\lambda}{2} \|w\|^{2}+\dsp\sum_{i=1}^{L}\xi_{i}
\\
\mbox{subject to} \quad 
& \dsp\min_{\| \Sigma_i^{-1/2}\Delta x_i\|_p\leq \gamma_i} y_{i}(\langle w,x_i+\Delta x_i\rangle +b)+\xi_{i} \geq 1, \quad 
& \forall i=1,\ldots,L,
\nonumber
\\
& \xi_{i} \geq 0,
& \forall i=1,\ldots,L.
\nonumber
\end{eqnarray}
Denoting $\|.\|_q$ the dual norm of $\|.\|_p$, we have 
$$\dsp\min_{\| \Sigma_i^{-1/2}\Delta x_i\|_p\leq \gamma_i} y_{i} \langle w,\Delta x_i\rangle
= 
-\gamma_i\left\| \Sigma_i^{\top/2} w\right\|_q.$$
Therefore, Problem~(\ref{ronoisySVM}) can be rewritten as
\begin{eqnarray}
\label{roSVM}
\dsp\min_{w,b,\xi} \quad   
& \frac{\lambda}{2} \|w\|^{2}+\dsp\sum_{i=1}^{L}\xi_{i}
\\
\mbox{subject to} \quad 
& y_{i}(\langle w,x_i\rangle+b)+\xi_{i}-\gamma_i\left\| \Sigma_i^{\top/2} w\right\|_q \geq 1, \quad 
& \forall i=1,\ldots,L,
\nonumber
\\
& \xi_{i} \geq 0,
& \forall i=1,\ldots,L.
\nonumber
\end{eqnarray}

\noindent This problem is a Second Order Cone Program (SOCP). There has been previous investigation on this formulation where the problem was solved using a primal-dual 
IPM~\cite{alizadeh2003} and solver packages such as~\cite{sedumi,mosek}. The complexity of primal-dual IPM is $O(s^3)$ operations per iteration and the number of iterations is $O(\sqrt{s}\times \log(1/\epsilon))$ where $\epsilon$ is the required precision and $s$ the number of cones in the problem. In Problem (\ref{roSVM}), the number of cones is $L+1$ ($L$ is the size of the dataset). Therefore these methods do not scale well as the size of the dataset increases. However, for small and medium size problems, 
they have given satisfying results~\cite{trafalis3,trafalis2}. For large problems, we propose alternative techniques in the next section.

\subsection*{Chance constrained approach}
Observe that if one has additional knowledge on the bound of the perturbations (ex: mean and variance of the noise distribution), one may want to consider the chance constrained approach rather than the worst case methodology above. Doing so, we can replace in Problem~(\ref{noisysvm}) 
the constraint $y_{i}(\langle w,x_i+\Delta x_i\rangle+b)+\xi_{i} \geq 1$ by
$$
\mathbb{P}_{\Delta x_i}\left[y_{i}(\langle w,x_i+\Delta x_i\rangle+b)+\xi_{i} \geq 1\right] \geq 1-\epsilon_i, 
\quad \forall i\in \{1,...,L\},
$$
where $\epsilon_i$ is a given positive confidence level. In other words, instead of ensuring the constraint for all realizations of $\Delta x_i$, we would like to ensure its probability of being satisfied to be greater than some threshold value $1-\epsilon_i$. Indeed, it was shown in~\cite{shivaswamy2006} that if $\Delta x_i$ follows a normal distribution $\mathcal{N}(0,\Sigma_i)$, the chance constraint problem can be formulated exactly as Problem~(\ref{roSVM}) if $p$ is taken as $2$ and $\epsilon_i$ such that $\gamma_i=\sqrt{\epsilon_i/(1-\epsilon_i)}$. The proof makes use of the multivariate Chebyshev inequality to replace the probability of ensuring the constraint by the Chebyshev bound. The two approaches are therefore related when $p=2$. In the work that is described next, we have chosen to exploit the worst case model as less assumptions are taken on the uncertainty set since $p$ can take any positive value greater than $1$ and we do not require explicit knowledge of variance of the perturbations.

\subsection*{Large scale robust framework}
\noindent We now consider the case where the dataset is very large and SOCP techniques cannot be used in practice. We reformulate the problem (\ref{roSVM}) as an equivalent unconstrained optimization problem:
\begin{equation}\label{rohinge}
\begin{array}{rll}
\dsp \min_{w,b}\quad f(w,b)=\frac{\lambda}{2}\|w\|^2 +  \dsp\sum_{i=1}^{L}\ell_{\Sigma_i}(y_i,\langle w,x_i\rangle+b),
\end{array}
\end{equation}
where $\ell_{\Sigma_i}$ is the robust loss function defined by
\begin{equation}\label{robustl}
\ell_{\Sigma_i}(y_i,\langle w,x_i\rangle+b)=\max\left\{0,1-y_i(\langle w,x_i\rangle+b) 
+ \gamma_i \| \Sigma_i^{\top/2} w\|_q\right\}.
\end{equation}

\noindent There are many techniques available to solve problem~(\ref{rohinge}). When $L$ is very large, stochastic approximation techniques may be considered as they have very low per-iteration complexity. For example, stochastic proximal methods have proven to be very effective for this type of large decomposable problems~\cite{villa2014}. The idea is to split the objective $f$ into two components $G(w,b)=g(w,b)+h(w)$ where
$$
g(w,b)=\dsp\sum_{l=1}^{L}g_l(w,b), 
\quad g_l(w,b)=\ell_{\Sigma_l}(y_l,\langle w,x_l\rangle+b), 
\quad h(w)=\frac{\lambda}{2}\|w\|^2,
$$
to perform a stochastic forward step on the decomposable component $g$ by taking a gradient approximation using only one random term $g_l$ of the decomposable sum, and finally to carry out a backward proximal step using $h$.

\subsection*{Controlling the robustness of the model}
Chance constraints and worst case robust formulations are based on a priori information on the noise bound or its variance. If these estimations are erroneous and over-estimated for example, the models may be too conservative. Additionally, worst case situations may be too pessimistics and may introduce additional non separability in the problem resulting in lower generalization performance of the model. This is the reason why we have proposed an alternative adaptive robust model that will have the advantage of being less conservative without taking any additional assumption on the probability distribution~\cite{CouellanWang2015}. The idea is to consider an adjustable subset of the support of the probability distribution of the uncertainties. Assuming an ellipsoidal uncertainty set, a reduced uncertainty set (reduced ellipsoid) is computed so as to minimize a generalization error. The model is referred to as \textit{safe SVM} rather than robust SVM. Mathematically, this is achieved by introducing a variable $\sigma$ defining a $\Sigma_{\sigma}$ matrix of a reduced ellipsoid where $\sigma$ is the vector of lengths of the ellipsoid along its axes. The resulting model can be cast as a bi-level program as follows:
\begin{equation}\label{bisvm}
\begin{array}{rll}
\dsp\min_{\sigma \in \mathbb{R}^n} \quad   & \dsp\sum_{j=1}^{N}\ell(y^v_j,\langle w^{*},x^v_j\rangle +b^{*})\\
\mbox{s.t.} \quad & \sigma_{min} \leq \sigma_t \leq \sigma_{max} \quad \forall t=1,\ldots,n\\
& (w^{*},b^{*})=\dsp\argmin_{w,b} \frac{\lambda}{2} \|w\|^{2}+\dsp\sum_{i=1}^{L}\ell_{\Sigma_{\sigma_t}}(y_i,\langle w,x_i\rangle +b)\\                        
\end{array}
\end{equation} 
where $(x^v_j,y^v_j)$, for $j=1,\ldots,N$, are the sample vectors and their labels from the validation fold. The upper and lower bounds $\sigma_{max}$ and $\sigma_{min}$ are parameters that control the minimum and maximum amount of uncertainty we would like to take into account in the model. They can be taken as rough bounds of the perturbations. The function $\ell$ is the standard SVM hinge loss function that will estimate the generalization error for one test sample and $\ell_{\Sigma_{\sigma}}$ is the robust loss function as defined in (\ref{robustl}). The matrix $\Sigma_{\sigma}$ is defined as $\left(\diag(\sigma)\right)^2$.\\
In~\cite{CouellanWang2015}, we have solved Problem~(\ref{bisvm}) with a stochastic bi-level gradient algorithm~\cite{CouellanWang2016}. Using a \textit{robust error} measure as an indicator of performance of the robustness of the model, we have shown that the proposed model achieves a better \textit{robust error} on several public datasets when compared to the SOCP formulation or formulation~(\ref{rohinge}). Using gaussian noise, we have also shown that the volume of the reduced ellipsoid computed by the technique was indeed increasing with variance of the noise, confirming that the model was auto adjusting its robustness to the noise in the data.    

%% file: NLRoSVM.tex
\section{Nonlinear robust SVM classification models}
\noindent There are situations where the model (\ref{primsvm}) will not be able to compute a sufficiently good linear separation between the classes if the data are highly non linearly separable. This phenomenon is actually even more present when noise is taken into account since robust models introduce additional non separability in the data. For this reason, it is interesting to investigate nonlinear robust formulations. Nonlinear SVM models are based on the use of kernel functions: 
\begin{eqnarray}
\label{mapprimsvm}
\dsp\min_{\alpha, b,\xi} \quad   
& \frac{\lambda}{2} \alpha^\top K \alpha+\dsp\sum_{i=1}^{L}\xi_{i}
\\
\mbox{subject to} \quad 
& y_{i}\left(\dsp\sum_{i=1}^{L} \alpha_jk(x_i,x_j)+b\right)+\xi_{i} \geq 1, \quad 
& \forall i=1,\ldots,L,
\nonumber
\\
& \xi_{i} \geq 0,
& \forall i=1,\ldots,L,
\nonumber
\end{eqnarray}
where $K$ is a kernel matrix whose elements are defined as 
$$k_{ij} = \langle\phi(x_i),\phi(x_j)\rangle = k(x_i,x_j)$$
and $k$ is a kernel function such as a polynomial, a Gaussian Radial Basis Function (RBF), or other specific choices of kernel functions~\cite{ShaweTaylor2004}.

\noindent One difficulty we are facing when designing robust counterparts formulations in the RKHS space is the understanding of how the uncertainty set is transformed through the implicit mapping chosen via the use of a specific kernel. The kernel function and its associated matrix $K$ are known but its underlying mapping $\phi$ is unknown. It is therefore difficult to seek bounds of the image of the uncertainty set in the original space through $\phi$. To address this issue, we investigate two methods : the \text{Random Fourier Features} (RFF) and the \text{Nystr\"{o}m method}. In both methods, the idea is to approximate the feature mapping $\phi$ by an explicit function $\tilde{\phi}$ from $\RR^n$ to $\RR^D$ ($n \leq D \leq L$) that will allow bounding of the image of uncertainties in the feature space. Specifically, if we can show that for all $i$ in $\{1,\ldots,L\}$ we can bound the uncertainty $\Delta\phi_i=\tilde{\phi}(x_i+\Delta x_i)-\tilde{\phi}(x_i)$ as follows:
\begin{equation}\label{bound}
\|R_i\Delta\phi_{i}\|_{\bar{p}} \leq \Gamma_i,
\end{equation}
for some $L_{\bar{p}}$ norm in $\RR^D$ and where $\Gamma_i$ is a constant, $R_i$ is some matrix, both depending on the choice of feature map approximation, we are able to formulate the nonlinear robust SVM problem as follows 

\begin{equation}\label{nlrosvm}
\begin{array}{rll}
\dsp \min_{\zeta \in \RR^D,b\in \RR}\quad \frac{\lambda}{2}\|\zeta\|^2 
+  \dsp\sum_{i=1}^{L}\max\left\{0,1-y_i(\langle\zeta,\tilde{\phi}(x_i)\rangle+b) 
+ \Gamma_i \|R_i^\top\zeta\|_{\bar{q}}\right\},
\end{array}
\end{equation}
\noindent where $\bar{q}$ is such that $\frac{1}{\bar{p}}+\frac{1}{\bar{q}}=1$. The construction of this problem is similar to the construction of problem~(\ref{roSVM}) in the linear case considering separation of approximate features $\tilde{\phi}(x_i)$ instead of data points $x_i$ and the use of a robust loss function in the feature space as in~(\ref{robustl}).

\noindent As far as we know, the model (\ref{nlrosvm}) is the first model that is specifically designed to perform nonlinear classification on large and uncertain datasets. The decomposable structure of the objective allows, as for the linear case, the use of  stochastic approximation methods such as the stochastic sub-gradient or stochastic proximal methods.\\
In the next 2 sections, we concentrate on deriving upper bounds of the form of~(\ref{bound}) by the RFF and the \Nys methods.

\input{RFF.tex}
\input{NYS.tex}

%% file: RFF.tex
\subsection{Bounding feature uncertainties via Random Fourier Features}\label{RFF}
\noindent We propose first to make use of the so-called \text{Random Fourier Features} (RFF) to approximate the inner product 
$\langle\phi(x_i),\phi(x_j)\rangle$ by $\tilde{\phi}(x_i)^\top \tilde{\phi}(x_j)$ in a low dimensional Euclidian space $\RR^D$, meaning that:
$$
k(x,y)=\langle\phi(x),\phi(y)\rangle\approx \tilde{\phi}(x)^\top \tilde{\phi}(y),\quad \forall x,y\in\RR^n.
$$
The explicit knowledge of the randomized feature map $\tilde{\phi}$ helps in bounding the uncertainties in the randomized feature space. Additionally, as we expect $D$ to be fairly low in practice, the technique should be able to handle large datasets as opposed to exact kernel methods that require the storage of a large and dense kernel matrix.

\noindent Mathematically, the technique relies on the Bochner theorem \cite{rahimi} which states that any shift invariant kernel $k(x,y)=k(x-y)$ is the inverse Fourier transform of a proper probability distribution $P$. If we define a complex-valued mapping $\hat{\phi}:\RR^n\rightarrow \mathbb{C}$ by $\hat{\phi}_{\omega}(x)=e^{j\omega ^\top x}$ and if we draw $\omega$ from $P$, the following holds:
$$
k(x-y)=\int_{\RR^n}P(\omega)e^{j\omega^\top (x-y)}d\omega=E\left[\hat{\phi}_{\omega}(x)\hat{\phi}_{\omega}(y)^*\right]
$$                      
where $^*$ denotes the complex conjugate. Furthermore, if instead we define 
$$\tilde{\phi}_{\omega,\nu}(x)=\sqrt{2}\left[\cos\left(\omega^\top x+\nu\right)\right],$$
where $\omega$ is drawn from $P$ and $\nu$ from the uniform distribution in $[0,2\pi]$, we obtain an approximate real-valued random Fourier feature mapping that satisfies
$$
E\left[\tilde{\phi}_{\omega,\nu}(x)\tilde{\phi}_{\omega,\nu}(y)\right]=k(x,y).
$$
The method for constructing RFF to approximate a Gaussian kernel could therefore be summarized as follows:
\begin{enumerate}
\item Compute the Fourier transform $P$ of the kernel $k$ by taking:
$$
P(\omega)=\frac{1}{2\pi}\int e^{-j\omega^\top \delta}k(\delta)d\delta; 
$$
\item Draw $D$ i.i.d. samples $\omega_1,\ldots,\omega_D$ in $\RR^n$ from $P$; 
\item Draw $D$ i.i.d. samples $\nu_1, \ldots, \nu_D$ in $\RR$ from the uniform distribution on~$[0,2\pi]$; 
\item The RFF $\tilde{\phi}_{\omega,\nu}(x)$ is given by
$$
\tilde{\phi}_{\omega,\nu}(x)=\sqrt{\dfrac{2}{D}}\left[\cos(\omega_1^\top x+\nu_1), \ldots, \cos(\omega_D^\top x+\nu_D)\right]^\top.
$$
\end{enumerate} 

\noindent In~\cite{SutherlandSchneider}, it has been shown that, when $k$ is Gaussian, the following choice of features leads to lower variance and should be preferred in practice:
$$
\tilde{\phi}_{\omega}(x)=\sqrt{\dfrac{2}{D}}\left[\cos(\omega_1^\top x), \sin(\omega_1^\top x), 
\ldots, 
\cos(\omega_{D/2}^\top x),\sin(\omega_{D/2}^\top x)\right]^\top.
$$

\begin{figure}
\caption{Random Fourier projections of uncertainties in a low dimensional feature space}
\label{rff}
\begin{center}
\begin{tabular}{cc}
\includegraphics[scale=0.27]{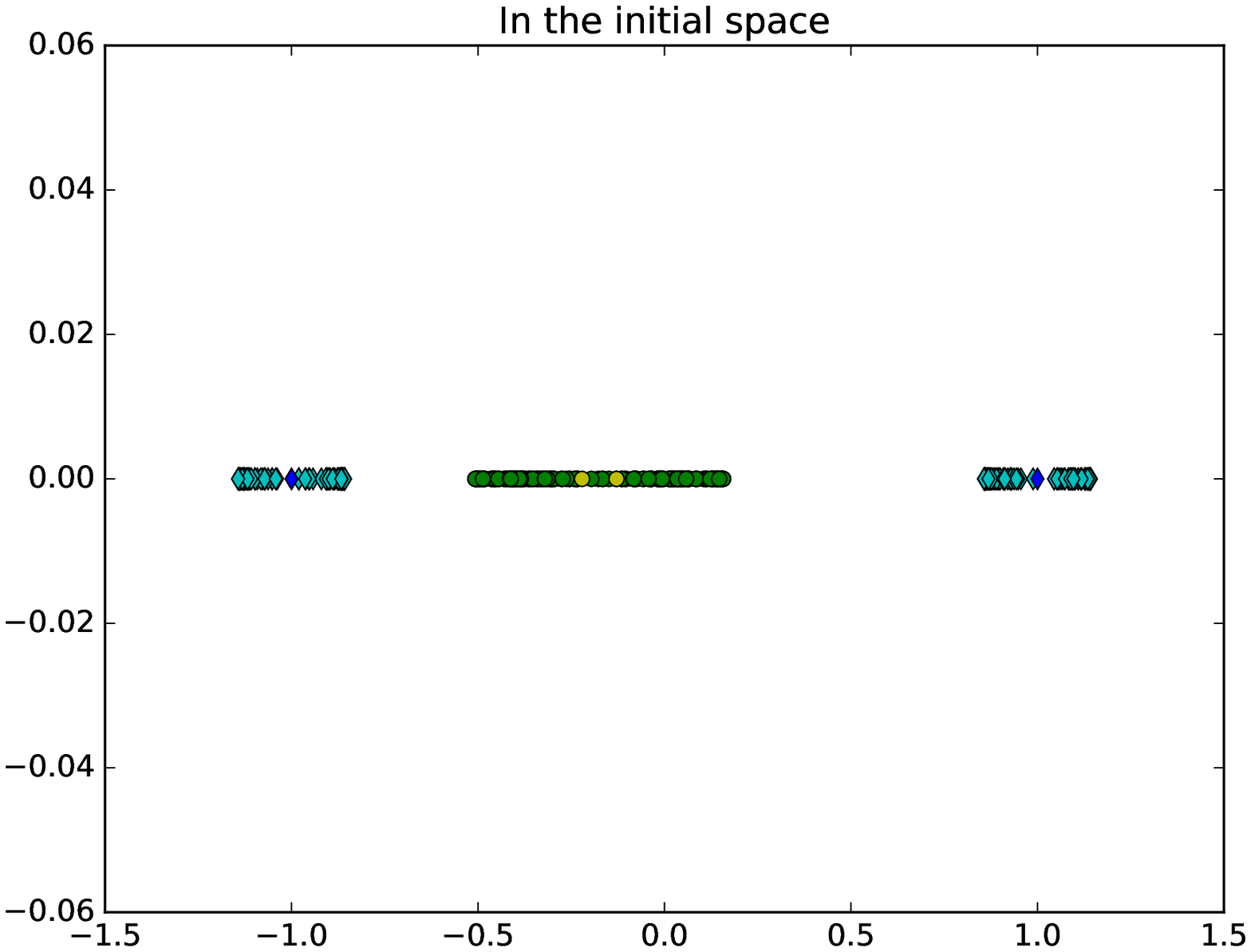}&
\includegraphics[scale=0.27]{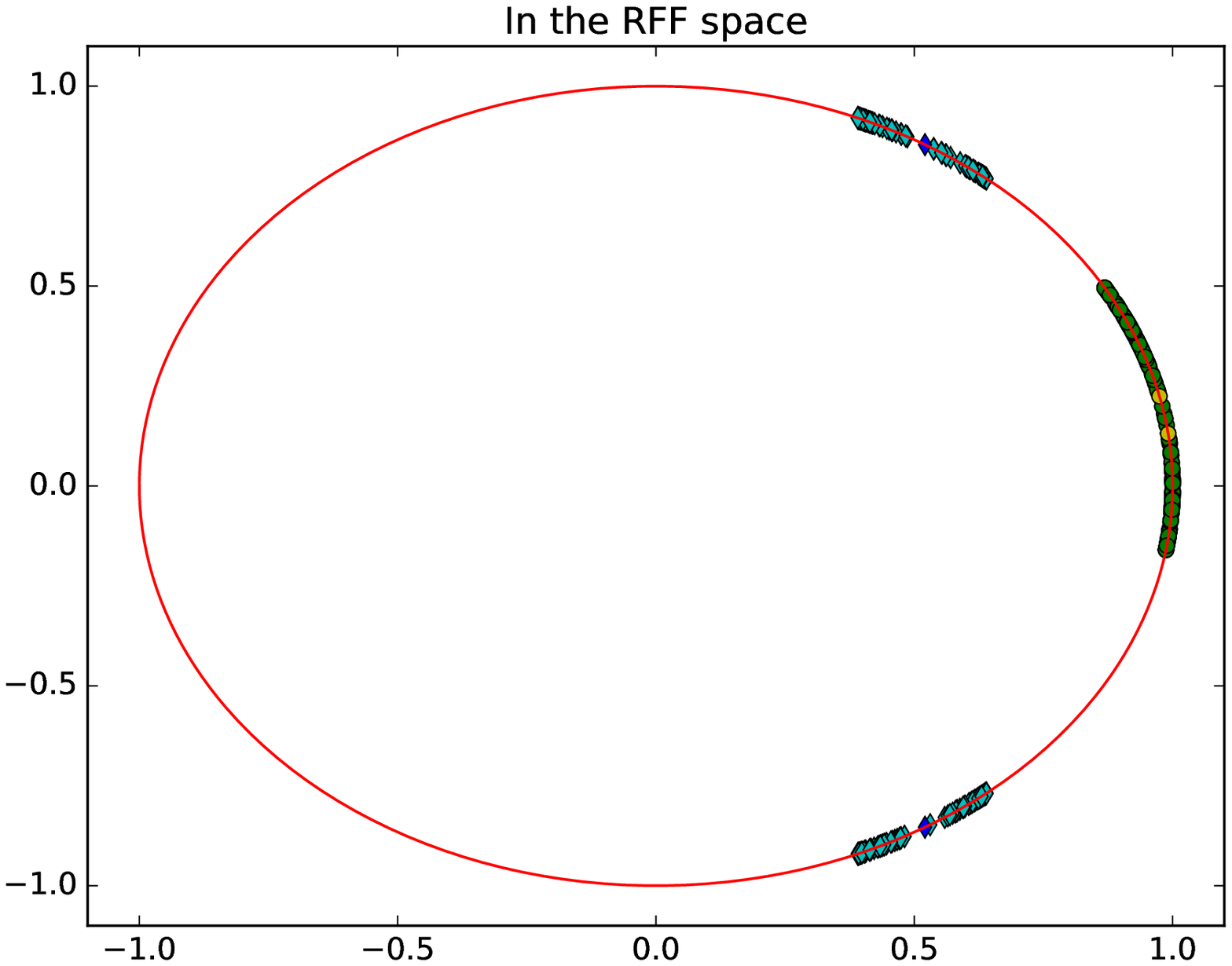}
\end{tabular}
\end{center}
\end{figure}

\noindent In Figure~\ref{rff}, a toy example is given. On the left side, one can see 4 nominal data points surrounded by their noisy observations. The 2 classes of points cannot be separated linearly. We construct random Fourier features with $D=2$ on the right side. One can observe that the noisy observations are now projected on the circle and linear separation is possible.

\noindent We are now interested in the analysis of the image of uncertainties when mapped in $\RR^D$ through $\tilde{\phi}_\omega$. 
With the explicit knowledge of $\tilde{\phi}_\omega$ 
and if we define $\Delta\phi_{\omega,i}=\tilde{\phi}_\omega(x_i+\Delta x_i)-\tilde{\phi}_\omega(x_i)$ 
as the induced uncertainty around each random feature in the randomized feature space $\RR^D$, we establish the following result:
\begin{proposition}
For all $i$ in $\{1,\ldots,L\}$, let $R_i$ be the block-diagonal matrix where the $D/2$ blocks of $R_i$ are rotation matrices of angle $-\omega_j^\top x_i$ for $j$ in $\{1,\ldots,D/2\}$. We have
$$
\|R_i\Delta\phi_{\omega,i}\|_{\bar{p}} \leq \Gamma_i,
$$
where $\Gamma_i$ are constants depending on $\gamma_i$, $\Sigma_i^{1/2}$ and the $D/2$ vectors $(\omega_i)$.
\end{proposition}

\noindent The proof of this result is as follows:

\bproof
Let $c_{ij} = \cos\Big(\omega_j^\top x_i\Big)$ and $s_{ij} = \sin\Big(\omega_j^\top x_i\Big)$. Using trigonometric relations, the $j$-th block of $\Delta\phi_{\omega,i}$ can be written
$$
\begin{array}{rcl}
(\Delta\phi_{\omega,i})_j
& = 
& \sqrt{\dfrac{2}{D}}\Bigg(\cos\Big(\omega_j^\top (x_i+\Delta x_i)\Big)-\cos\Big(\omega_j^\top x_i\Big),
\\
&
& \hspace*{2cm}            \sin\Big(\omega_j^\top (x_i+\Delta x_i)\Big)-\sin\Big(\omega_j^\top x_i\Big)\Bigg)^\top
\\
& = 
& \sqrt{\dfrac{2}{D}}\Bigg(c_{ij}\cos\Big(\omega_j^\top \Delta x_i\Big)
-s_{ij}\sin\Big(\omega_j^\top \Delta x_i\Big)-c_{ij},
\\
&
& \hspace*{2cm} s_{ij}\cos\Big(\omega_j^\top \Delta x_i\Big)
+c_{ij}\sin\Big(\omega_j^\top \Delta x_i\Big)-s_{ij},\Bigg)^\top
\\
& = 
& \sqrt{\dfrac{2}{D}} P_{ij}
\left(
\begin{array}{c}
\cos\Big(\omega_j^\top \Delta x_i\Big)-1
\\
\sin\Big(\omega_j^\top \Delta x_i\Big)
\end{array}
\right)
\end{array}
$$
where $$
P_{ij} = 
\left(
\begin{array}{rr}
c_{ij} & -s_{ij}
\\
s_{ij} & c_{ij}
\end{array}
\right)
$$
is the rotation matrix of angle $\omega_j^\top x_i$. Consequently, we obtain that
$$
P_{ij}^\top (\Delta\phi_{\omega,i})_j = \sqrt{\dfrac{2}{D}} 
\left(
\begin{array}{c}
\cos\Big(\omega_j^\top \Delta x_i\Big)-1
\\
\sin\Big(\omega_j^\top \Delta x_i\Big)
\end{array}
\right).
$$

\noindent Using the following bounding schemes
$$
|\cos(\theta) -1| \leq \dfrac{\theta^2}{2}, \quad |\sin(\theta)| \leq |\theta|,
$$
we have
\begin{equation}\label{cc}
\begin{array}{rcl}
\left|\cos\Big(\omega_j^\top \Delta x_i\Big)-1\right| 
& \leq 
& \dfrac{\left(\omega_j^\top \Delta x_i\right)^2}{2}
\\[0.2cm]
\left|\sin\Big(\omega_j^\top \Delta x_i\Big)\right| 
& \leq 
& \left| \omega_j^\top \Delta x_i\right|.
\end{array}
\end{equation}


\noindent Using H\"{o}lder inequality, we can write
\begin{equation}\label{bb}
\begin{array}{rcl}
\left|\omega_j^\top \Delta x_i\right| 
& = 
& \left|\omega_j^\top \Sigma_i^{1/2}\Sigma_i^{-1/2}\Delta x_i\right|
\\[0.2cm]
& \leq 
& \left\|\omega_j^\top \Sigma_i^{1/2}\right\|_q  \left\|\Sigma_i^{-1/2}\Delta x_i\right\|_p
\\
& \leq 
& \left\| \Sigma_i^{\top/2} \omega_j\right\|_q  \gamma_i,
\end{array}
\end{equation}
and 
$$
\begin{array}{rcl}
\left|\cos\Big(\omega_j^\top \Delta x_i\Big)-1\right| 
& \leq 
& \min\left(2,\dfrac{\left(\gamma_i\left\| \Sigma_i^{\top/2} \omega_j\right\|_q\right)^2}{2}\right)
\\
\left|\sin\Big(\omega_j^\top \Delta x_i\Big)\right| 
& \leq 
& \min\left(1,\gamma_i\left\| \Sigma_i^{\top/2} \omega_j\right\|_q\right).
\end{array}
$$
\noindent Introducing the notations 
$$R_i = {\rm diag}(P_{i,1}^\top, ..., P_{i,D/2}^\top),$$  
$$\alpha_{i,j} := 
\min\left(2,\dfrac{\left(\gamma_i\left\| \Sigma_i^{\top/2} \omega_j\right\|_q\right)^2}{2}\right)$$
and 
$$\beta_{i,j} := 
\min\left(1,\gamma_i\left\| \Sigma_i^{\top/2} \omega_j\right\|_q\right),$$ 
we finally obtain the expected bound
$$
\left\|R_i \Delta\Phi_{\omega,i}\right\|_{\bar{p}} \leq \Gamma_i
$$
where
$$\Gamma_i := \left\{
\begin{array}{rl}
\sqrt{\dfrac{2}{D}} \dsp\sum_{j=1}^{D/2} (\alpha_{i,j}+\beta_{i,j}) & \mbox{if } \bar{p} = 1 ;
 \\
\sqrt{\dfrac{4}{D}\dsp\sum_{j=1}^{D/2} \alpha_{i,j}} & \mbox{if } \bar{p} = 2 ;
 \\
\sqrt{\dfrac{2}{D}} \max\left(
\dsp\max_{j=1, ..., D/2} \alpha_{i,j},
\dsp\max_{j=1, ..., D/2} \beta_{i,j}
\right) & \mbox{if } \bar{p} = \infty.
\end{array}
\right.
$$

\eproof

%% file: NYS.tex
\subsection{Bounding feature uncertainties via the \Nys method}

\noindent We now investigate another method. Note that in the following, we will use the notation $\|.\|$ for the euclidean norm $\|.\|_2$. When working with large datasets, the computation and storage of the entire kernel matrix $K$ whose size is $L^2$ becomes difficult. For this reason, the \Nys method has been proposed to compute a low rank approximation of the kernel matrix~\cite{gittens_mahoney_16a,trokicic16,Homrighausen16,LiLuKwokLu15,Choromanska13,Chang13}. 
We can summarize the method as follows:
\begin{enumerate}
\item Choose an integer $m$ between the number of attributes $n$ and the size $L$ of the dataset.
\item Choose $m$ vectors $(\xh_i)_{i = 1, ..., m}$ among the $L$ training vectors.
\item Build the associated kernel matrix $\Kh \in \mathbb{R}^{m\times m}$ defined by
$$\Kh_{ij} := k(\xh_i,\xh_j).$$ 
\item Perform an eigenvalue decomposition of this matrix $\Kh$ to obtain
\begin{itemize}
\item its rank $r$,
\item a $m\times r$ matrix $\Usr := \left(\us{1}, ..., \us{r} \right)$ 
such that $\left(\Usr\right)^\top \Usr$ is the identity matrix, 
\item and a diagonal matrix $\Lsr := \diag(\ls{1}, ..., \ls{r})$
\end{itemize}
such that $\Kh \Usr = \Usr \Lsr$.
\item The \Nys $\phish(x)$ can be chosen as follows :
$$
\phish(x)= \left(\Lsr\right)^{-1/2}\left(\Usr\right)^\top \ksh(x),
$$
with
$$
\ksh(x) = \left(\ks(x,\xh_1), \ks(x,\xh_2), ..., \ks(x,\xh_m)\right)^\top.
$$
\end{enumerate}

\noindent As before, we are interested by the analysis of the uncertainties when they are mapped to the approximate feature space induced by the \Nys method through  $\phish$. We define $\Delta\phi_{\sigma,i}=\phish(x_i+\Dxi)-\phish(x_i)$ 
as the uncertainty in the feature space $\RR^r$ and show the following result:

\begin{proposition} 
For all $i$ in $\{1,\ldots,L\}$, let $R_i = \left(\Lsr\right)^{1/2}$, we have
$$
\|R_i\Delta\phi_{\sigma,i}\|_2 \leq \Gamma_i,
$$
where $\Gamma_i$ are constants depending on $\gamma_i$, $\Sigma_i^{1/2}$, $\sigma$ and the $m$ vectors $(\xh_i)_{i = 1, ..., m}$. 
\end{proposition}

\bproof
\noindent Using the expression of the feature mapping constructed via the \Nys method, we have
\begin{eqnarray*}
\left(\Lsr\right)^{1/2} \left(\phish(x_i+\Dxi)-\phish(x_i)\right)
& = 
& \left(\Usr\right)^\top \left(\ksh(x_i+\Dxi)-\ksh(x_i)\right)
\\
& = 
& \left(
\begin{array}{c}
\us{1}^\top \left(\ksh(x_i+\Dxi)-\ksh(x_i)\right)
\\
\vdots
\\
\us{r}^\top \left(\ksh(x_i+\Dxi)-\ksh(x_i)\right)
\end{array}
\right).
\end{eqnarray*}

\noindent Consequently, we can write
\begin{eqnarray*}
\left\| \left(\Lsr\right)^{1/2} \left(\phish(x_i+\Dxi)-\phish(x_i)\right)\right\|_2^2
& = 
& \sum_{j=1}^r \left(\us{j}^\top \left(\ksh(x_i+\Dxi)-\ksh(x_i)\right)\right)^2
\\
& \leq
& \sum_{j=1}^r \left\|\ksh(x_i+\Dxi)-\ksh(x_i)\right\|_2^2
\\
& = 
& r \left\|\ksh(x_i+\Dxi)-\ksh(x_i)\right\|_2^2.
\end{eqnarray*}

\noindent We are now interested in computing an upper bound of the left hand side of the above expression. Since
\begin{eqnarray*}
\ksh(x_i+\Dxi)-\ksh(x_i)
& = 
& \left(
\begin{array}{c}
\ks(x_i+\Dxi,\xh_1)- \ks(x_i,\xh_1)
\\
\vdots
\\
\ks(x_i+\Dxi,\xh_m)-\ks(x_i,\xh_m)
\end{array}
\right)
\end{eqnarray*}
we get 
\begin{eqnarray*}
\left\|\ksh(x_i+\Dxi)-\ksh(x_i)\right\|_2^2
& = 
& \sum_{j=1}^m \left(\ks(x_i+\Dxi,\xh_j)-\ks(x_i,\xh_j)\right)^2.
\end{eqnarray*}

\noindent If the uncertainties are defined with the choice of Euclidian norm (meaning $p=q=2$), if we also define the kernel $\ks$ with the Euclidian norm (meaning $\ks(x,z) := e^{-\dfrac{\|x-z\|_2^2}{2\sigma^2}}$), and observing that

\begin{eqnarray*}
\ks(x_i+\Dxi,\xh_j)-\ks(x_i,\xh_j)
& = 
& e^{-\dfrac{\|x_i+\Dxi-\xh_j\|^2}{2\sigma^2}}-e^{-\dfrac{\|x_i-\xh_j\|^2}{2\sigma^2}}
\\
& = 
& e^{-\dfrac{\|x_i-\xh_j\|^2}{2\sigma^2}}
\left(
e^{-\dfrac{2\Dxi^\top (x_i-\xh_j)+\|\Dxi\|^2}{2\sigma^2}}
-1
\right),
\end{eqnarray*}
\noindent we can obtain
\begin{eqnarray*}
\left(\ks(x_i+\Dxi,\xh_j)-\ks(x_i,\xh_j)\right)^2
& = 
& \left(e^{-\dfrac{\|x_i-\xh_j\|^2}{2\sigma^2}}\right)^2
\left(e^{-\dfrac{2\Dxi^\top (x_i-\xh_j)+\|\Dxi\|^2}{2\sigma^2}}\right)^2
\\
& - 2
& \left(e^{-\dfrac{\|x_i-\xh_j\|^2}{2\sigma^2}}\right)^2
e^{-\dfrac{2\Dxi^\top (x_i-\xh_j)+\|\Dxi\|^2}{2\sigma^2}}
\\
& + 
& \left(e^{-\dfrac{\|x_i-\xh_j\|^2}{2\sigma^2}}\right)^2
\\
& = 
& \left(e^{-\dfrac{\|x_i-\xh_j\|^2}{2\sigma^2}}\right)^2
\left(e^{-\dfrac{2\Dxi^\top (x_i-\xh_j)}{2\sigma^2}}\right)^2
\left(e^{-\dfrac{\|\Dxi\|^2}{2\sigma^2}}\right)^2
\\
& - 2
& \left(e^{-\dfrac{\|x_i-\xh_j\|^2}{2\sigma^2}}\right)^2
e^{-\dfrac{2\Dxi^\top (x_i-\xh_j)}{2\sigma^2}}
e^{-\dfrac{\|\Dxi\|^2}{2\sigma^2}}
\\
& + 
& \left(e^{-\dfrac{\|x_i-\xh_j\|^2}{2\sigma^2}}\right)^2.
\end{eqnarray*}

\noindent We have easily $\left(e^{-\dfrac{\|\Dxi\|^2}{2\sigma^2}}\right)^2 \leq 1$. We are now aiming at bounding from below the terms $e^{-\dfrac{\|\Dxi\|^2}{2\sigma^2}}$ and $e^{-\dfrac{2\Dxi^\top (x_i-\xh_j)}{2\sigma^2}}$. Using $\|\Sigma_i^{-1/2}\Delta x_i\|_2\leq \gamma_i$, we can write
$$
0\leq \|\Delta x_i\| 
= \|\Sigma_i^{1/2}\Sigma_i^{-1/2}\Delta x_i\|
\leq \|\Sigma_i^{1/2}\| \|\Sigma_i^{-1/2}\Delta x_i\|
\leq \gamma_i \|\Sigma_i^{1/2}\|.
$$
\noindent which consequently leads to, $-\gamma_i^2 \|\Sigma_i^{1/2}\|^2 \leq -\|\Delta x_i\|^2 \leq 0$ and
$$
e^{-\dfrac{\gamma_i^2 \|\Sigma_i^{1/2}\|^2}{2\sigma^2}} \leq e^{-\dfrac{\|\Dxi\|^2}{2\sigma^2}} \leq 1.
$$

\noindent Now, writing 
$\Dxi^\top (x_i-\xh_j) 
= (x_i-\xh_j)^\top \Sigma_i^{1/2}\Sigma_i^{-1/2}\Delta x_i$, 
\noindent we have
$\left| \Dxi^\top (x_i-\xh_j) \right| 
\leq \gamma_i \|\Sigma_i^{\top/2} (x_i-\xh_j)\|$
\noindent and thus
$$
e^{-\dfrac{2\gamma_i \|\Sigma_i^{\top/2} (x_i-\xh_j)\|}{2\sigma^2}}
\leq
e^{-\dfrac{2\Dxi^\top (x_i-\xh_j)}{2\sigma^2}}
\leq
e^{\dfrac{2\gamma_i \|\Sigma_i^{\top/2} (x_i-\xh_j)\|}{2\sigma^2}}.
$$

\noindent Using these bounds, we can write 
\begin{eqnarray*}
\left(\ks(x_i+\Dxi,\xh_j)-\ks(x_i,\xh_j)\right)^2
& \leq
& \left(e^{-\dfrac{\|x_i-\xh_j\|^2}{2\sigma^2}}\right)^2
\left(e^{\dfrac{2\gamma_i \|\Sigma_i^{\top/2} (x_i-\xh_j)\|}{2\sigma^2}}\right)^2
\\
& - 2
& \left(e^{-\dfrac{\|x_i-\xh_j\|^2}{2\sigma^2}}\right)^2
e^{-\dfrac{2\gamma_i \|\Sigma_i^{\top/2} (x_i-\xh_j)\|}{2\sigma^2}}
e^{-\dfrac{\gamma_i^2 \|\Sigma_i^{1/2}\|^2}{2\sigma^2}}
\\
& + 
& \left(e^{-\dfrac{\|x_i-\xh_j\|^2}{2\sigma^2}}\right)^2
\end{eqnarray*}

\noindent and come back to 

\begin{eqnarray*}
\left\|\ksh(x_i+\Dxi)-\ksh(x_i)\right\|_2^2
& \leq 
& \sum_{j=1}^m \left(e^{-\dfrac{\|x_i-\xh_j\|^2}{2\sigma^2}}\right)^2
\left(
1+
\left(e^{\dfrac{2\gamma_i \|\Sigma_i^{\top/2} (x_i-\xh_j)\|}{2\sigma^2}}\right)^2
\right)
\\
& - 
& 2e^{-\dfrac{\gamma_i^2 \|\Sigma_i^{1/2}\|^2}{2\sigma^2}}
\sum_{j=1}^m \left(e^{-\dfrac{\|x_i-\xh_j\|^2}{2\sigma^2}}\right)^2
e^{-\dfrac{2\gamma_i \|\Sigma_i^{\top/2} (x_i-\xh_j)\|}{2\sigma^2}}
\end{eqnarray*}
\noindent or
$$
\left\|\ksh(x_i+\Dxi)-\ksh(x_i)\right\|_2^2
\leq
\sum_{j=1}^m k_{ij}^2 \left(\dfrac{1}{\tau_{ij}^2}+1\right) 
-2\rho_i \sum_{j=1}^m k_{ij}^2 \tau_{ij}
$$
\noindent where we have denoted
$$
k_{ij} = e^{-\dfrac{\|x_i-\xh_j\|^2}{2\sigma^2}}, \quad
\tau_{ij} = e^{-\dfrac{2\gamma_i \|\Sigma_i^{\top/2} (x_i-\xh_j)\|}{2\sigma^2}}, \quad
\rho_i = e^{-\dfrac{\gamma_i^2 \|\Sigma_i^{1/2}\|^2}{2\sigma^2}}.
$$

\noindent Finally, we have obtained
\begin{eqnarray*}
\left\| \left(\Lsr\right)^{1/2} \left(\phish(x_i+\Dxi)-\phish(x_i)\right)\right\|_2^2
& \leq
& \Gamma_i^2
\end{eqnarray*}
\noindent with
$$
\Gamma_i = \sqrt{r \left( 
\sum_{j=1}^m k_{ij}^2 \left(\dfrac{1}{\tau_{ij}^2}+1\right) 
-2\rho_i \sum_{j=1}^m k_{ij}^2 \tau_{ij} 
\right)}.
$$

\eproof

%% file: commentary.tex
\section{A word on the tightness of bounding schemes}

We would like here to analyze under which conditions the bounds we have proposed are tight. Ensuring some tightness will avoid over estimating the uncertainties in the approximate feature space that could make class separation more difficult.\\
Consider the bounding scheme from Section \ref{RFF} and where $k$ is chosen as the Gaussian kernel, meaning that $\ks(x,z) := e^{-\dfrac{\|x-z\|^2}{2\sigma^2}}$. The method relies on the inequalities (\ref{cc}) that ensure that

$$
\begin{array}{rcl}
\left|\cos\Big(\omega_j^\top \Delta x_i\Big)-1\right| 
& \leq 
& \dfrac{\left(\omega_j^\top \Delta x_i\right)^2}{2}
\\
\left|\sin\Big(\omega_j^\top \Delta x_i\Big)\right| 
& \leq 
& \omega_j^\top \Delta x_i.
\end{array}
$$
These bounds are rather tight if the angle $\left|\omega_j^\top \Delta x_i\right|$ is smaller than some given $\theta_{max}$. To ensure (with high probability) that the angle will remain below $\theta_{max}$ we state and proof the following result:

\begin{proposition}\label{prop}
Assume that $\Sigma_i^{1/2}=\Sigma^{1/2}$ where $\Sigma^{1/2}$ is a constant diagonal matrix. When $q=2$, if 
$$
\sigma \geq \frac{3\gamma_i\|\Sigma^{1/2}\|_F}{\theta_{max}},
$$
with high probability (greater than $0.997$), we have $\left|\omega_j^\top \Delta x_i\right| \leq \theta_{max}$.
\end{proposition}

\bproof
Remember that $\omega_j$ are distributed according to the Fourier transform of the kernel $k$, therefore we have $\omega_j \sim \mathcal{N}\left(0,\frac{1}{\sigma^2}\right)$ 
and we know that $\omega_j \in \left[ -\frac{3}{\sigma},\frac{3}{\sigma}\right]$ with high probability (greater than $0.997$). Additionally, we have shown in (\ref{bb}) that for all $j$ in $\{1,\ldots,D/2\}$,
\begin{equation}\label{dd}
\left|\omega_j^\top \Delta x_i\right| \leq \left\| \Sigma_i^{\top/2} \omega_j\right\|_q  \gamma_i.
\end{equation}
When $q=2$ and $\Sigma_i^{1/2}=\Sigma^{1/2}=\diag{(s)}$, we have
$$
\left\| \Sigma_i^{\top/2} \omega_j\right\|_2^2  = \dsp\sum_{k=1}^{n}(s_k\omega_{jk})^2 \leq  \frac{9\|\Sigma^{1/2}\|_F^2}{ \sigma^2},
$$
and therefore, when we take 
\begin{equation}\label{ee}
\frac{9 \gamma_i^2 \|\Sigma^{1/2}\|_F^2}{\sigma^2} \leq \theta_{max}^2,
\end{equation}
using (\ref{dd}) we have, with high probability, that $\left|\omega_j^\top \Delta x_i\right| \leq \theta_{max}$.
The inequality (\ref{ee}) can also be written as $\sigma \geq \frac{3\gamma_i\|\Sigma^{1/2}\|_F}{\theta_{max}}$.\eproof \\

The result in Proposition~\ref{prop} actually says that there is a trade-off between bounding the uncertainties in the approximate feature space and separating correctly the uncertain features. For the latter, we would like to achieve relatively small (but not too small) values of $\sigma$. On the other hand to ensure tightness of our bounding scheme, the RFF method requires that $\sigma$ should be taken sufficiently large. Therefore, the right trade-off for choosing the $\sigma$ value is depending on data. The condition in Proposition \ref{prop} should then be incorporated in the definition of the $\sigma$-grid that one usually defines when designing a cross-validation procedure in practice.